\documentclass[sigconf]{acmart}

\usepackage{multirow}

\AtBeginDocument{%
  \providecommand\BibTeX{{%
    \normalfont B\kern-0.5em{\scshape i\kern-0.25em b}\kern-0.8em\TeX}}}

\begin{document}

\title{AdnFM: An Attentive DenseNet based Factorization Machine for Click-Through-Rate Prediction}

\author{Kai Wang}
\affiliation{%
  \institution{Tencent}
  \country{China}
  }
  
\email{wangjinjie722@gmail.com}

\author{Chunxu Shen}
\affiliation{%
  \institution{Tencent}
  \country{China}
  }
\email{lineshen@tencent.com}

\author{Chaoyun Zhang}
\affiliation{%
  \institution{Tencent Lightspeed \& Quantum Studios}
  \country{China}
  }
\email{vyokkyzhang@tencent.com}

\author{Wenye Ma}
\affiliation{%
  \institution{Tencent}
  \country{China}
  }
\email{mawenye@gmail.com}


\begin{abstract}
    In this paper, we introduce a novel deep learning-based model named AdnFM, to attack the Click-Through-Rate (CTR) prediction problem. The AdnFM includes two important components to extract both low-order and high-order features of users and items, to jointly learn a comprehensive representation for the prediction. It further combines residual learning and an attention mechanism, to enable high-order features interactions and weight their importance dynamically. We conduct extensive experiments to evaluate the performance of the proposed architecture. Results show that the AdnFM outperforms popular baselines on two offline dataset. We deploy our model on an online CTR prediction application. Online A/B test demonstrates that the proposed AdnFM achieves remarkable performance and significantly outperform other benchmarks. 
\end{abstract}




\maketitle
\pagestyle{plain}
\section{\textbf{troduction}}
The Click-Through-Rate (CTR) prediction is a core task in a recommender system. It has been widely employed in a range of applications, such as online advertising and news ranking \cite{graepel2010web}. Due to the natural quality of the CTR prediction, designing a recommender system with high efficiency and low complexity is not straightforward. As the features extracted from users are usually sparse, heterogeneous and high-dimensional, the size of the input to a recommender system becomes non-negligible \cite{eirinaki2018recommender}. Working with such large scale of feature set will significantly increase the complexity of the system.

To mitigate this issue, many researches seek to reduce the complexity from the model perspective. For example, generalized linear models \cite{thai2010recommender} become a popular method for the CTR problem due to its simplicity. However, by reason of its poor representability, heavy feature engineering is usually required to ensure the performance. Another popular model is the factorization machine (FM) \cite{RendleICDM}. It is employed to capture the second-order pairwise feature interactions by a low-rank matrix in a factorized form. Compared to linear models, the FM provides more meaningful feature interaction with acceptable complexity. It has been widely deployed in industry and its variants achieve state-of-the-art performance in various applications \cite{JuanWWW,LuWSDM,NguyenSIGIR,RendleSIGIR,ZhongCIKM,rafm}.



\begin{figure*}[htbp]
\centering
\includegraphics[height=2.5in]{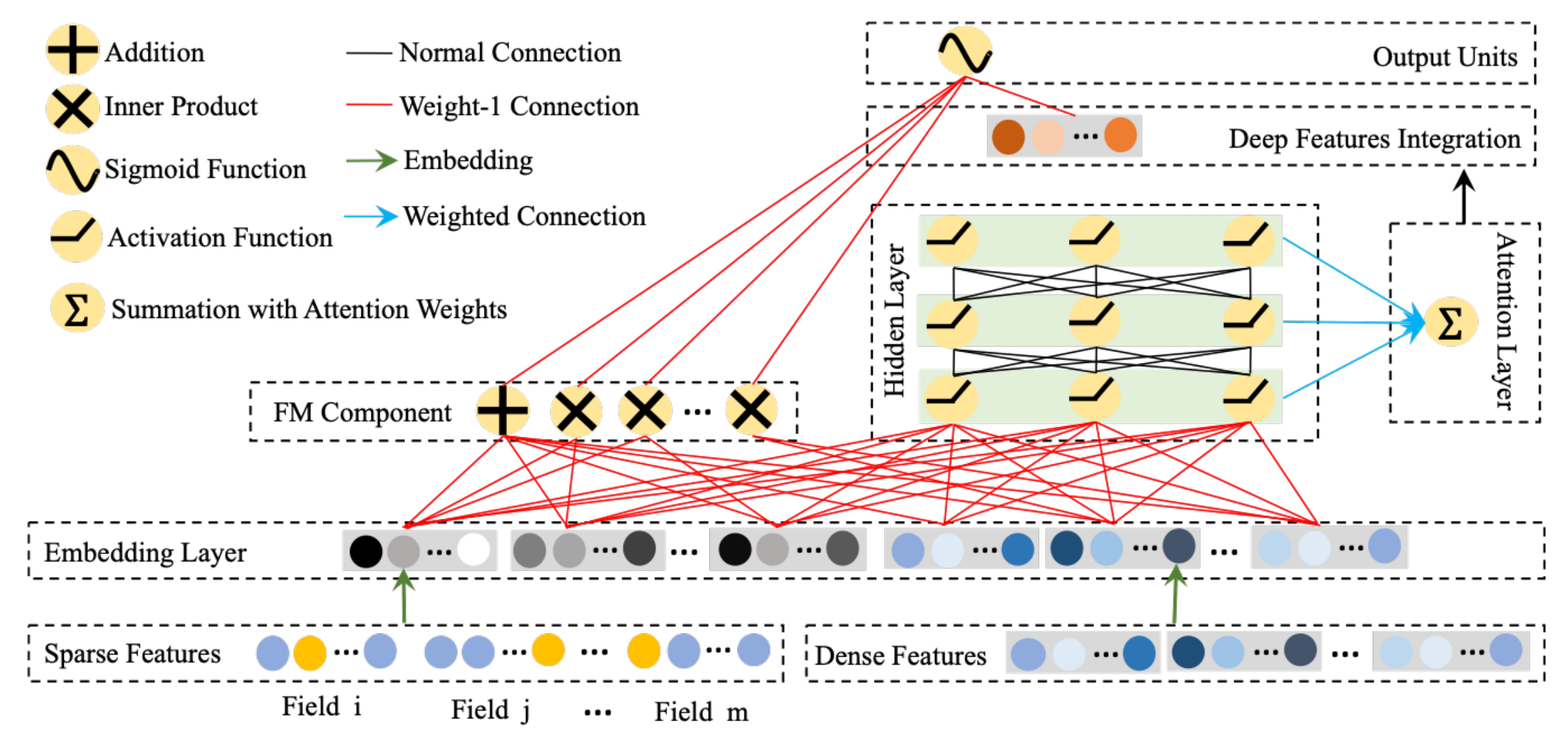}
\caption{The overall structure of our proposed AdnFM. The AdnFM includes two main components: an Adn component and a FM component. The Adn component learns the hierarchical attention weighted between different layers through an attention neural network. Its outputs of high-order features are integrated via a weighted pooling layer. }
\label{model} 
\end{figure*}

Underpinned by the advance parallel computing, deep neural networks (DNNs) have achieved remarkable performance in different fields \cite{voulodimos2018deep, socher2012deep, zhang2019deep}. This inspires researchers to employ the deep learning method to attack the CTR prediction problem. Well-designed DNN-base models bring significant improvement to CTR prediction by its remarkable ability of feature extraction. 
For example, Deep \& Cross Network \cite{deepandcross} applies a multi-layer residual structure to learn explicit high-order representation of features. Attentional Factorization Machine (AFM) \cite{AFM} employs the attention mechanism to automatically learn the weights of interacted features. The xDeepFM \cite{xdeepfm} models feature interactions by a novel Compressed Interaction Network (CIN) component. These deep learning based solutions bring the CTR system to a higher level.
ome are explicit. 

In industrial deployment, recommander systems usually employ first-order (e.g. age, gender) and second-order (e.g. combinations of age and gender, age and heights etc.) as input for the CTR prediction. This only captures the superficial aspects of the users and items. Nevertheless, it is recognized that high-order features can also be combined to construct high-dimensional and non-linear features \cite{neuralFM}.  While high-order features are normally neglected, they includes more specific and richer users' information, which can contribute to the performance of the system. However, generating high-order features usually requires substantial computing resource due to its high complexity \cite{blondel2016higher}. In addition, it is difficult to find the most important high-order features, as there exist numerous feature combinations. 

In this paper, we propose a novel model named Attentive DenseNet based factorization machine (AdnFM) to attack the CTR prediction problem. The original DenseNet \cite{densenet} is an popular CNN based architecture, which was first introduced in the computer vision field. Compared to traditional CNNs, the residual learning in DenseNet alleviates vanishing-gradient problems, strengthens the feature propagation, encourages feature reusing, and reduces the model complexity. This makes the model particularly suitable for the CTR prediction problem, as feature reusing can share more mutual information between users and items. Inspired by the DenseNet and NFM \cite{neuralFM}, we use the hidden layers in the DNN to represent high-order features, and integrate them through a weighted pooling layer. Feature interactions are further strengthened by the residual paths in the dense skip-connection. In addition, we employ a FM component to serve for the second-order feature interaction with remarkable efficiency. By jointly training a factorization machine and a deep neural network, high-order features are generated automatically. We subsequently employ the attention mechanism to integrate these features, by weighting their importance automatically. This enables to weight and capture the feature from different orders and perspectives, which contributes to the overall performance.
Overall, the contributions of the paper are summarized as follow:

\begin{itemize}
\item  We introduce a novel deep learning based architecture named AdnFM to attack the CTR prediction problem. The AdnFM uses hidden layers to represent high-order features, which is generated automatically and efficiently via jointly training with a FM. 
\item The AdaFM employs the residual learning to enable feature interactions. The hierarchical weights of high-order features are learnt through a level-based attention mechanism. A weighted pooling layer is then utilized to integrate features extracted.

\item We conduct extensive experiments on two real-world public dataset. Results demonstrate the effectiveness of the proposed model on CTR prediction and recommendation problems, as it achieves up to 0.6\% higher CTR compared to DeepFM. 
\item We deploy our model for the CTR prediction on a popular Tencent mobile game.  Online A/B tests demonstrate that the proposed model outperforms different baseline methods by up to 12\% in terms of the CTR.
\end{itemize}
In what follows, we introduce the background of the CTR problem, as a preamble. 


\begin{figure*}[htbp] 
    \centering 
    \includegraphics[height=3.0in]{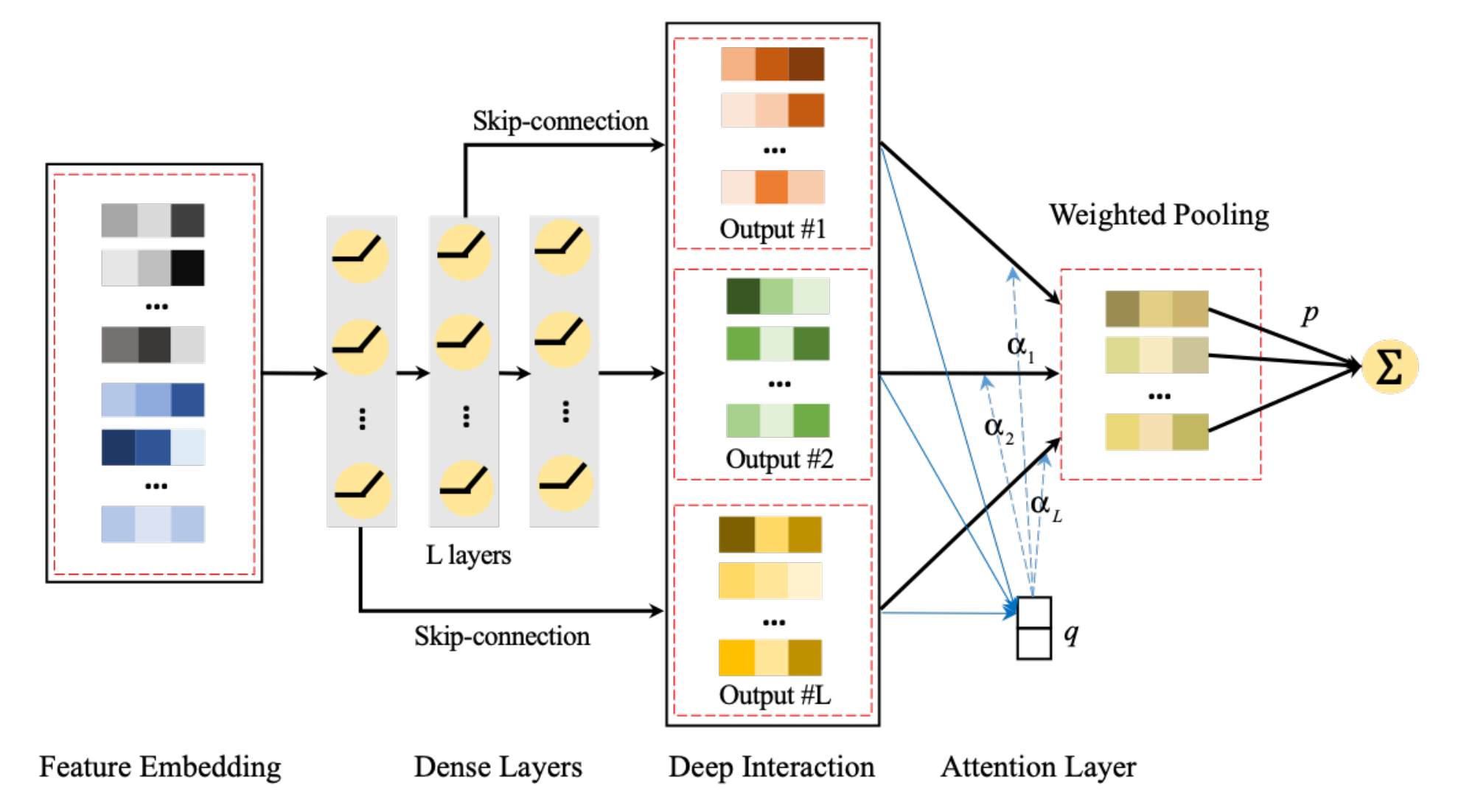} 
    \caption{The overall structure of the Adn.} 
    \label{Adn} 
\end{figure*}

\section{\textbf{CTR Prediction in a Nutshell}}
The CTR Prediction  is one of the most critical and profitable task in recommender systems for Internet companies, as very small improvement of CTR would bring huge profits for the company. However, raising CTR by even a small margin is not straightforward. As there are usually tens of thousands of features fed into the model, and the amount of click data samples can be very large, it is difficult to extract useful feature for the prediction.


The input of the recommender system consist of user-item vector pair $(u, i)$. Specifically, the  vector $u$ denotes user's features such as age, gender, etc. 
The item vector $i$ represents an item's features. For example, in news recommendation scenario, the $i$ could contain title, category, etc. Feature engineering techniques, such as feature discretization and crossing, are usually employed to transform an user-item pairs $(u, i)$ into a feature vector $\mathbf{x}=[x_1, x_2, \ldots, x_D]$ (or sample). 
Since the feature engineering will expand the dimensions of features with one-hot operations, the feature vector $\mathbf{x}$ is usually sparse and high-dimensional. Given a feature vector, the recommender system calculates an item's click probability $p$ by the user $u$. The system is then updated by the user's real behaviors (whether he/she actually clicks the item). Therefore, the CTR prediction problem can be modeled as predicting a probability using $(u, i)$ based on the feature vector $x$:


\begin{equation}
    p = P(\mathit{click}|u,i) = P(\mathit{click}|x).
\end{equation}
$p$ can be considered as the probability that a user $u$ clicks an item $i$. The label $y \in \{0,1\}$, indicates whether the user actually clicks ($y=1$) the item or not ($y=0$). The key idea is to model the prediction as a function of feature vector $x$. The function is usually parameterized by learning variables $\theta$:
\begin{equation}
    p = p_\theta(x).
\end{equation}
We define a loss function to measure the difference between $p$ and the true label $y$. In the CTR prediction, we can use the logistic loss function:
\begin{equation}
    f(\theta) = -ylog(p_\theta(x)) - (1-y)log(1-p_\theta(x)).
\end{equation}
By minimizing the loss function from a training dataset, the model is trained to optimize its parameters set $\theta$, and predicts the CTR for items.

\section{\textbf{Proposed Method}}
In this section, we introduce our proposed Attentive DenseNet based Factorization Machine (AdnFM). AdnFM incorporates both low-order features from traditional FM and high-order features extracted from a DenseNet-alike architecture. The overall structure of AdnFM is shown in Figure \ref{model}. Specifically, the AdnFM extracts hierarchical features and strengthens feature interactions via an attention network with residual path. We named such a component as Adn. The hierarchical features are subsequently  processed by weighted pooling layer. Finally, the AdnFM combines the output from FM with Adn to perform prediction. 

\subsection{\textbf{Sparse Input and the Embedding Layer}}
Data in CTR prediction tasks have two types of values, namely numerical and categorical. For example, the price of the item is numerical, e.g. [price=12.3]. Numerical values are usually discretized and represented by one-hot vectors. Another type of feature is grouped categorical. For example, the gender of a user [gender=Male] is categorical. A categorical value can be directly transformed to an one-hot binary vector. As the one-hot vector will expand the feature space, the processed inputs is a high-dimensional and sparse representation. We denote $\mathbf{x} \in \mathbb{R}^D$ as the sparse feature vector and $D$ is the dimension of the feature space. We use $y$ to denote the label, which indicates users' behaviors. For CTR prediction, we assume $y\in\{0, 1\}$, which represents whether a user clicks the given item or not. The CTR prediction can be also formulated as a regression problem, where $y$ is a real number to represent the score that a user rates on the given item in this case.

Embedding is a common technique to transform high-dimensional sparse features into low-dimensional dense vectors. The output of the embedding layer can be represented as a $K\times F$ matrix, where each column is an embedding vector, $K$ is the dimension of embedding layer, and $F$ is the number of feature fields. We represent the embedding as: 
\begin{equation}
    E = [\mathbf{e}_1, \mathbf{e}_2, \ldots, \mathbf{e}_F],
\end{equation}
where $\mathbf{e}_i\in\mathbb{R}^K$ is the embedding vector. The embedding dimension $K$ is a hyper-parameter that can be tuned in the model.

\subsection{\textbf{The FM Component}}
The FM component in AdnFM is a vanilla factorization machine proposed in \cite{RendleICDM}. The FM captures the first-order features as well as the second-order pairwise feature interactions by low-rank matrices in a factorized form. It only needs $O(M\cdot D)$ space, where $M$ is a hyper-parameter representing the rank and $D$ is the dimension of the feature vectors. The general form of FM is defined as:
\begin{equation}
y_{FM} = b + \langle\mathbf{w}, \mathbf{x}\rangle + \sum^D_{i=1}\sum^{D}_{j=i+1}  \langle\mathbf{v}_i,\mathbf{v}_j\rangle x_i x_j,\\
\end{equation}
where $b$ is the bias term, $\mathbf{w}$ is the linear component and $\mathbf{v}_i\in\mathbb{R}^M$ (with $i=1, 2, \ldots, D$) are the variables used to capture the pairwise feature interactions.  $M$ is normally much smaller than $D$.
In our proposed model, we adopt the same method in \cite{guo2017deepfm} and reuse the parameters in FM as the embedding layer for jointly training. We set $K=M$ and
\begin{equation}
    \mathbf{e}_i = \sum_{l\in\mbox{Field}_i}\mathbf{v}_l x_l,
\end{equation}
where $\mbox{Field}_i$ denotes the index set of feature field $i$.

\subsection{\textbf{The Adn Component}}
Our proposed method employs a deep neural network as the automatic feature extractor, which enables to generate high-order features without human interference. Normally, we expect outputs of deeper layer in a model can represent higher-order features. However, the plain structures of traditional CTR prediction methods ignore the relations between layers, which hinders the feature interactions. Inspired by the DenseNet, we create shorter connections between different layers in the neural network and model the importance between different orders of features dynamically by an Adn component. We show its structure in Figure \ref{Adn}. The dense and deep interaction layer concatenates high-order features represented by hidden units. The attention layer weights the importance between hidden layers. Then the high-order features will be integrated in a weighted-pooling layer.

\subsubsection{Dense Layer}
The input of the dense layer is passed from the embedding layer, as defined above. We concatenate the embedding vectors to construct the input:
\begin{equation}
    H_0 = \mbox{concat}(\mathbf{e}_1, \mathbf{e}_2, \ldots, \mathbf{e}_F).
\end{equation}
Recent work has shown that DNNs can be substantially deeper, more accurate and efficient to train with the help of skip-connections. Therefore, we adopt a feed-forward neural network with all the hidden layers in the same size. Let $H_k$ with $k=1, 2, \ldots, L$ denotes the different hidden layers, we have $H_k\in\mathbb{R}^d$, where $d$ is the dimension. The structure of neural network can be represented as:
\begin{subequations}
\begin{equation}
    H_1 = \mbox{ReLU}(W^{(0)} H_0 + b^{(0)}),
\end{equation}
\begin{equation}
    H_{k+1} = \mbox{ReLU}(W^{(k)} H_k + b^{(k)}).
\end{equation}
\end{subequations}

\subsubsection{The Deep Interaction}
In analogy to the DenseNet, we extract output from all dense layers, and concatenate those as the final output. 
The DenseNet \cite{densenet} was first proposed in the computer vision field, and has achieved remarkable performance. It strengthens hierarchical feature integration by encouraging feature interactions. Inspired by the DenseNet, we introduce similar residual learning in our Adn. We treat the values of hidden layers as high-order features, which contain more comprehensive information compared to low-order counterparts. As the model goes deeper, the features are interacted through the skip-connections. This is more cost-efficient compared to traditional high-order feature interaction methods.  The skip-connection at the end of the layer achieves efficient integration of all features. Eventually, the output of L-layer dense network can be formally writen as: 
\begin{equation}
    Out_{L} = [H_1, H_2, \dots, H_L],
\end{equation}
where $L$ is depth of the network. The output of deep interaction layer is subsequently fed into an attention layer.

\subsubsection{The Attention Layer}
Although feature interactions in the deep interaction layer can benefit the performance, the importance of different features are not equal, as some features are naturally more important than others. To weight each feature dynamically, we employ an attention mechanism to balance hierarchical features from different layers. The attention layers can be defined as:
\begin{subequations}\label{att}
\begin{equation}
    \alpha_{k}' = \langle h, \mbox{ReLU}(W_a H_{k} + b_a)\rangle,
\end{equation}
\begin{equation}
    \alpha_{k} = \frac{\exp(\alpha_{k}')}{\sum_{k=1}^{L} \exp(\alpha_{k}')}\label{aij}.
\end{equation}
\end{subequations}
where $W_a \in \mathbb{R}^{d \times e}$, $d$ is the number of hidden units of each layer, $b_a \in \mathbb{R}^{e}$, $h \in \mathbb{R}^{e}$ and $e$ denote the number of hidden units in the attention network. A softmax function is used to normalize the attention score, and ReLUs function are used as the activation function \cite{AFM}. 

\subsubsection{Weighted Pooling}
To achieve high-order feature integration, we use a weighted pooling as the final operation combined with the attention mechanism to process the output from the attention layer.  Therefore, the final logits of the Adn becomes the inner-product of a learning variable and the weighted sum of all hierarchical features, defined as follow:
\begin{equation}
   y_{Adn} = \langle q, \sum_{k=1}^{L}\alpha_{k}H_{k}\rangle.
\end{equation}
In the equation, $q \in \mathbb{R}^{d}$ is a learnable variable, and $\alpha$ represents the hierarchical relations between different layers.

\subsection{\textbf{Prediction Layer}}
Finally, outputs from the FM component and the Adn are be combined in the prediction layer:
\begin{equation}
    p = \mbox{Pred}(y_{FM} + y_{Adn}),
\end{equation}
where $\mbox{Pred}$ is the prediction function. In the CTR prediction task, we use the Sigmoid function for this purpose:
\begin{equation}
    \mbox{Pred}(a) = \sigma(a) =\frac{1}{1+ \exp(-a)} \label{sigmoid}.
\end{equation}
The overall structure can the AdnFM can be optimized by minimizing the following crossed-entropy loss function:
\begin{equation}
    Loss_{CTR}(y, p) = -y \log(p) - (1-y)\log(1-p) \label{logloss}.
\end{equation}
In addition, it is worth mentioning that the AdnFM model can be applied to the regression problem, where we can set the prediction as the identity function and use the mean square error. 



\section{\textbf{Experiments}}
We conduct extensive experiments to evaluate the performance of our proposed AdnFM model on two public datasets, and compared with popular baselines in the industry.

\subsection{\textbf{Experiment Setup}}
\subsubsection{Datasets}
We evaluate our proposed method on two popular machine learning tasks: CTR prediction, and regression for recommendation. For CTR prediction, we adopt the Sigmoid function as defined in (\ref{sigmoid}) and the crossed-entropy function defined in (\ref{logloss}) as the last layer and the loss function. For this task, we use the Criteo dataset \footnote{https://www.kaggle.com/c/criteo-display-ad-challenge}, which contains click log of an online advertisement service collected in 7 days. The dataset contains over 45 million samples, each of which embraces 13 integer features and 26 categorical features.

For the regression task, we adopt the mean square loss defined in (\ref{identity}) and (\ref{mseloss}). We use MovieLens (20M) datasets \footnote{https://grouplens.org/datasets/movielens/} for this task. The MovieLens contains anonymous ratings (score from 0 to 5) for 27,278 movies from 128,493 users. For this dataset, we use user ID, movie ID, and movie genres as raw features. For both datasets, we randomly separate the data into training (80\%), validation (10\%), and test (10\%) set.

\subsubsection{Metrics}
For the CTR prediction task, we utilize Area Under Curve (AUC) and the crossed-entropy loss to evaluate the performance of all methods.  While we use Root Mean Square Error (RMSE) for the evaluation for the regression task.

\subsubsection{Baselines}
We compare our proposed approach with popular baselines used in the industry. These methods are considered as efficient, with low complexity and easy to deploy. These include: 
\begin{itemize}
\item \textbf{LR} Linear Regression;
\item \textbf{FM} Factorization Machines \cite{RendleICDM};
\item \textbf{AFM} Attentional Factorization Machines \cite{AFM};
\item \textbf{DNN} Multi-layer perceptron. In our experiments, DNN's structure is the same as deep component in Wide\&Deep;
\item \textbf{Wide\&Deep} Wide and Deep \cite{wideanddeep};
\item \textbf{DeepFM} Deep Factorization Machines \cite{guo2017deepfm};
\end{itemize}



\subsection{\textbf{Offline Performance Evaluation}}

\begin{table}
\centering
\caption{Performance evaluation on the Criteo and MovieLens Datasets.}
\begin{tabular}{lrrr}
\toprule
\ & 
\multicolumn{2}{c}{Criteo}{\centering} & MovieLens \\
\midrule
Model & AUC & LogLoss  & RMSE    \\
\midrule
LR          & 0.7736  & 0.4732   & 0.8844  \\
FM          & 0.7752  & 0.4721   & 0.8676  \\
AFM         & 0.7436  & 0.4934   & 0.8884  \\
DNN         & 0.7846  & 0.4648   & 0.8607  \\
Wide\&Deep  & 0.7878  & 0.4613   & 0.8597  \\
DeepFM      & 0.7858  & 0.4629   & 0.8614  \\
AdnFM       & \textbf{0.7919}  & \textbf{0.4576}   & \textbf{0.8513}  \\
\bottomrule
\end{tabular}

\label{result}
\end{table}

We compare the proposed AdnFM method with popular baselines on both Criteo and MovieLens datasets. Considering the feature number difference between Criteo and MovieLens, we set the number of hidden units in Adn structure to $\{128, 128, 128\}$ (3 layers) for the Criteo and $\{128, 128\}$ (2 layers) for the MovieLens respectively. These setting makes good trade-off between accuracy and computation cost. Table \ref{result} shows the evaluation summary on both dataset. Observe that the proposed AdnFM outperforms others models consistently on both datasets. Specifically, the AdnFM outperforms DeepFM by 0.77\% for AUC in Criteo dataset and 1.18\% for RMSE in MovieLens dataset for all setting. This makes significant improvement for existing approaches.


\subsection{\textbf{The Effect of the Attention Mechanism}}

\begin{table}
\centering
\caption{Performance evaluation on the attention mechanism. The first number in the size column denotes the width of all layers and the second number represents the depth.}
\begin{tabular}{lrrrr}
\toprule
\ & \ &
\multicolumn{2}{c}{Criteo}{\centering} & MovieLens\\
\midrule
Model & Size & AUC & LogLoss &  RMSE\\
\midrule
DenseFM   & \multirow{2}{*}{64 $\times$ 2} & 0.7849  & 0.4634  &  0.8582\\
AdnFM       &  & \textbf{0.7884}  & \textbf{0.4610}  &  \textbf{0.8551}\\
\midrule
DenseFM   &\multirow{2}{*}{64 $\times$ 3} &  0.7854 & 0.4630   & 0.8605\\
AdnFM       &       & \textbf{0.7888}  &  \textbf{0.4606}   & \textbf{0.8595}\\
\midrule
DenseFM   &\multirow{2}{*}{128 $\times$ 2} & 0.7875  & 0.4613   & 0.8579 \\
AdnFM       &           & \textbf{0.7914}  & \textbf{0.4581} & \textbf{0.8513}\\
\midrule
DenseFM   &\multirow{2}{*}{128 $\times$ 3} & 0.7868  & 0.4618  & 0.8571  \\
AdnFM       &           & \textbf{0.7919}  & \textbf{0.4576}  & \textbf{0.8538} \\
\bottomrule
\end{tabular}

\label{ablation}
\end{table}

\begin{figure}[htbp]
    \centering 
    \includegraphics[height=2.5in]{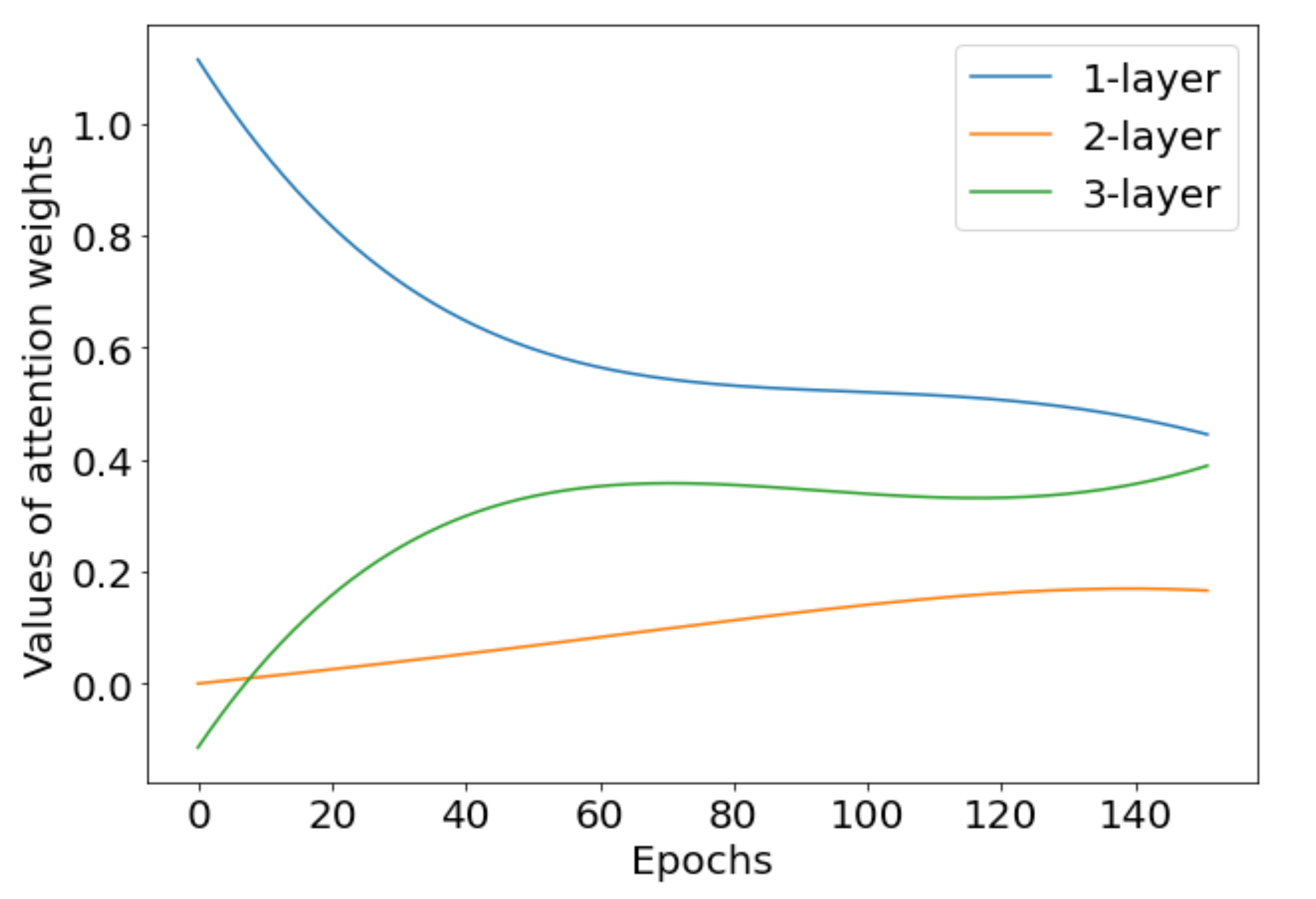} 
    \caption{The evolution of the attention weights of dense layers during training.} 
    \label{weight} 
\end{figure}

In order to validate the effectiveness of the attention network in our model, we visualize the evolution of the attention weights of dense layers during training on the Criteo datasets in Figure-\ref{weight}. Observe that the weight of the first layer drops during the training, while the weights of the second and third layer increase. As features represented by deeper layers are expected to be more abstract and higher-order, results in Figure \ref{weight} demonstrates that higher-order features are becoming increasingly important during training and therefore are given more attention.

In addition, we perform ablation study for the attention mechanism to evaluate its effectiveness. To this end, we remove the attention network and use the concatenation of the hidden layers as output to re-evaluate its performance. We name the modified model as DenseFM, and show the result comparison in Table \ref{ablation}. Observe that the AdnFM outperforms the DenseFM consistently across all model structures and all performance metrics. This proves the attention mechanism is significant in the model, as it can weight the importance of feature dynamically.

\subsection{\textbf{Online Performance Evaluation}}
\begin{figure}[htbp]
    \centering 
    \includegraphics[height=2.5in]{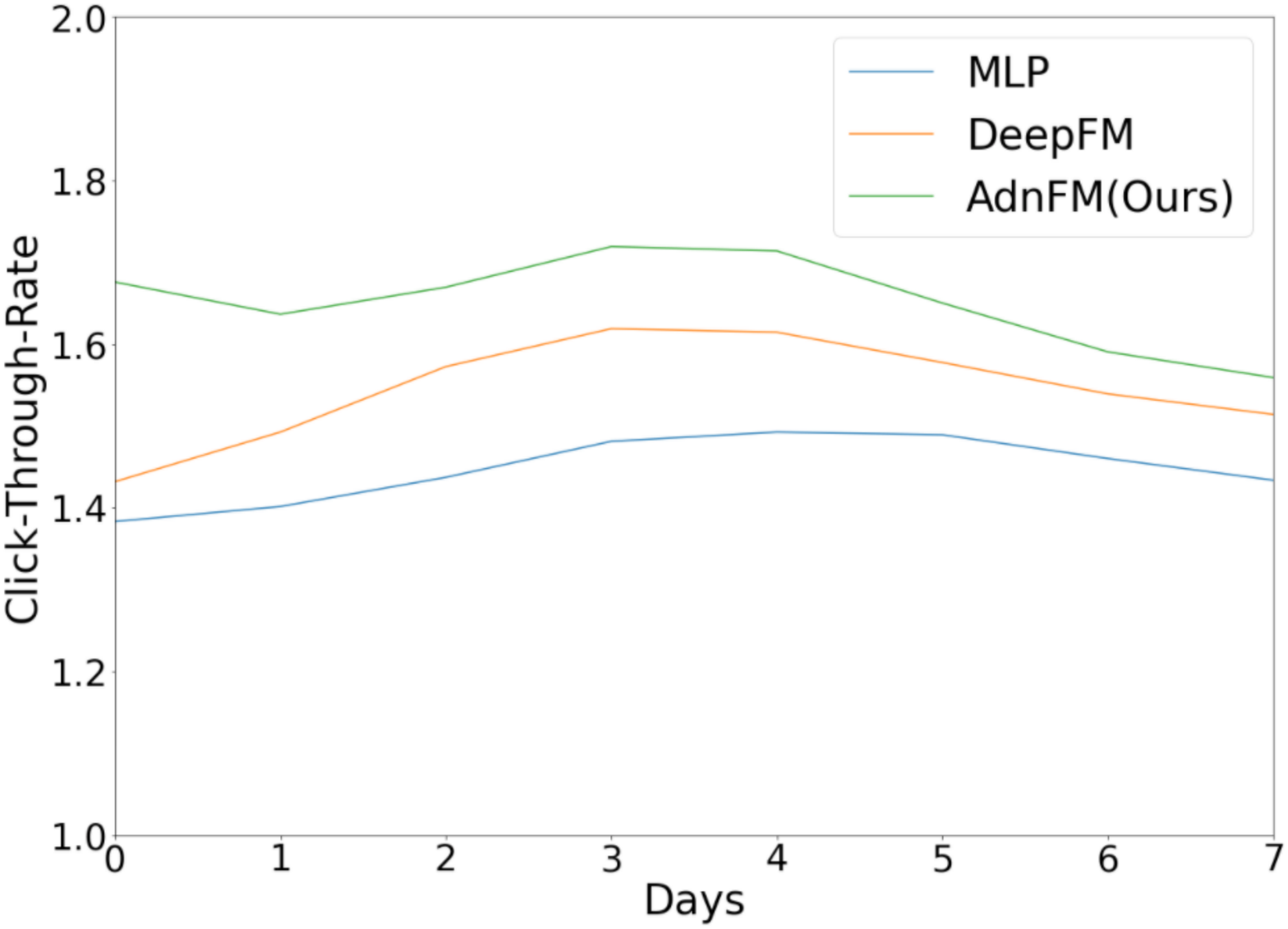} 
    \caption{Online A/B test results in Tencent game honor of kings.} 
    \label{online} 
\end{figure}

Finally, we deploy the proposed model in a Tencent's online scenario for the advertisement CTR prediction task of a popular mobile game Honor of Kings. Honor of Kings is one of the most popular MOBA game in the market in China, thus conducting A/B test in its player community is very convincing. The model is employed in the ranking phase of the advertisement system of player community, to predict whether the advertisement will be clicked by user. We conduct online A/B test for one week with averagely partitioned data stream. The number of unique user is over 10,000,000 and page view is over 1,000,000,000. As we can see in Figure \ref{online}, our proposed Adnfm significantly outperforms other two baselines MLP and DeepFM, as it achieve 6\% and 12\% higher CTR. Noted that there existing daily fluctuation on CTR, which is resulted by the activities and stream difference between weekday and weekend.

Overall, our proposed AdnFM achieves consistently good performance on both offline and online evaluation. Due to its remarkable performance in A/B test, the proposed architecture has become a important component in a Tencent's advertisement CTR prediction application.

\section{\textbf{Related Work}}

\subsection{\textbf{Deep Learning-Based Models}}

Several deep learning-based models have been proposed for the CTR prediction. Wide\&Deep \cite{wideanddeep} jointly trains a wide model and a deep model, which leverages the effectiveness of feature engineering and learns implicit feature interactions. DeepFM \cite{guo2017deepfm} introduces a FM layer as a wide component and uses a deep component to learn implicit feature interactions. Factorization Machine supported Neural Network (FNN) \cite{FNN} uses pre-trained factorization machines for field embedding to learn high-order feature interactions. Instead of using pairwise feature interactions, the Deep \& cross network \cite{deepandcross} captures feature interactions of bounded degrees in an explicit fashion. Another FiBiNet proposed in \cite{fibinet} effectively learns the feature interactions via the bilinear function in the Squeeze-and-Excitation block \cite{hu2018squeeze}.

Several research demonstrates that using users' historical behavior data can improve the prediction performance. The YoutubeNet \cite{youtubenet} transforms embeddings of users' watching lists into a vector of fixed length by an average pooling. The Deep Interest Network (DIN) \cite{DIN} uses the attention mechanism to learn the representation of the users' historical behaviors with respect to the target item. The Deep Interest Evolution Network (DIEN) \cite{DIEN} uses auxiliary loss and adjusts the expression of current behavior to next behavior, and then models the specific interest evolving process for different target items with the structure called GRU with Attention Update Gate (AUGRU).

\subsection{\textbf{The Attention Mechanism}}
The attention mechanism enables neural networks to weight the importance of different features, and has achieved impressive results in natural language processing \cite{vaswani2017attention}. Enlightened by the transformer model, researchers bring attention mechanism into CTR prediction and facilitate feature selection automatically, such as AFM \cite{AFM}, FiBiNet \cite{fibinet}, DIN \cite{DIN} and AutoInt \cite{song2019autoint}. The AFM places the attention in the interaction section in FM, but it remains restricted to the high time complexity when performing high-order interactions. The FiBiNet pays more attention to embedding layer with a self-designed SENET layer. Usually, DNN-based models with the attention mechanism outperform their baselines without attention substantially.

\section{\textbf{Conclusion}}
In this paper, we propose a novel deep learning-based model AdnFM  for the CTR prediction problem. The AdnFM jointly 
trains a FM and the Adn component to extracted both low-order and high-order features. Inspired by the DenseNet, we employ residual learning to enable feature interactions in an efficient way. Features from different levels are weighted dynamically through an attention layer. We conduct extensive experiments on both offline evaluation and online applications. The AdnFM outperforms other popular baselines on two offline CTR prediction dataset. Online A/B test on a Tencent's CTR prediction application shows that the AdnFM improve the traditional DeepFM and MLP by up to 12\%. It has becomes an important component in the Tencent's online application due to its remarkable performance.

\bibliographystyle{ACM-Reference-Format}
\bibliography{acmart}

\appendix

\end{document}